\documentclass{article}

\usepackage[preprint,nonatbib]{neurips_2020}

\usepackage[utf8]{inputenc} % allow utf-8 input
\usepackage[T1]{fontenc}    % use 8-bit T1 fonts
\usepackage{hyperref}       % hyperlinks
\usepackage{url}            % simple URL typesetting
\usepackage{booktabs}       % professional-quality tables
\usepackage{amsfonts}       % blackboard math symbols
\usepackage{nicefrac}       % compact symbols for 1/2, etc.
\usepackage{microtype}      % microtypography

\usepackage{graphicx}
\usepackage{makecell}
\usepackage{amsmath}
\usepackage{xcolor}

\title{Assessing Intelligence in Artificial Neural Networks}

\author{%
  Nicholas J. Schaub \\
  National Center for the Advancement of Translational Sciences \\
  National Institutes of Health\\
  \texttt{nick.schaub@nih.gov} \\
  \\
  Axle Informatics \\
  \texttt{nick.schaub@axleinfo.com} \\
\And
  Nathan Hotaling \\
  National Center for the Advancement of Translational Sciences \\
  National Institutes of Health\\
  \texttt{nathan.hotaling@nih.gov} \\
  \\
  Axle Informatics \\
  \texttt{nathan.hotaling@axleinfo.com} \\
}

\begin{document}

\maketitle

\begin{abstract}
The purpose of this work was to develop of metrics to assess network architectures that balance neural network size and task performance. To this end, the concept of neural efficiency is introduced to measure neural layer utilization, and a second metric called artificial intelligence quotient (aIQ) was created to balance neural network performance and neural network efficiency. To study aIQ and neural efficiency, two simple neural networks were trained on MNIST: a fully connected network (LeNet-300-100) and a convolutional neural network (LeNet-5). The LeNet-5 network with the highest aIQ was 2.32\% less accurate but contained 30,912 times fewer parameters than the highest accuracy network. Both batch normalization and dropout layers were found to increase neural efficiency. Finally, high aIQ networks are shown to be memorization and overtraining resistant, capable of learning proper digit classification with an accuracy of 92.51\% even when 75\% of the class labels are randomized. These results demonstrate the utility of aIQ and neural efficiency as metrics for balancing network performance and size.
\end{abstract}

\section{Background}

There is strong motivation to create large artificial neural networks (ANNs) to increase task performance, and there are efforts to develop new methods of training networks containing up to trillions of parameters.\cite{rajbh2019zero} However, there are a number of challenges to training and using large ANNs. Some of these challenges have found solutions in new types of layers or architectural designs. For example, some networks are too large for the task they are being trained to perform, resulting in poor task performance relative to a smaller network trained to perform the same task. This issue is resolved with skip or residual connections, which may require depth rescaling using another neuron layer.\cite{he2015deep} Another major issue is overtraining/memorization, a common problem in nearly all modern neural networks. Overtraining causes ANNs to memorize inputs rather than learn a set of rules that generalize to new data, and it has been shown that modern networks are highly capable of memorizing randomized image labels.\cite{zhang2016understanding} To combat overtraining and memorization, a variety of techniques and layers are used, including data augmentation,\cite{hernandez2018} dropout layers,\cite{srivastava14} and even the creation of new neural networks in the case of adversarial neural network training.\cite{antoniou2017data} Thus, the drive to create larger ANNs is often compounded by further increases in ANN size by using architecture components designed to combat the problems of large networks, which make the deployment of these networks at scale a challenge.

The current state of ANN research has an analagous mindset to the human intelligence research prior to the 1990s. The general opinion in human intelligence research was that an individual with high intelligence was more capable of performing tasks with high performance because his or her brain would be more capable of recruiting large numbers of neurons. This changed in the 1990s with multiple publications from Haier \textit{et al},\cite{haier1988,haier1992b} one of which showed individuals with high intelligence had higher scores on Tetris but had lower brain metabolism while playing the game.\cite{haier1992a} This finding led to the formulation of the neural efficiency hypothesis, which states that a key factor to intelligence is the capacity of the brain to perform a task by using the smallest amount of neural activity.\cite{neubauer2009} In the context of human intelligence and neural efficiency, the current drive to increase neural network task accuracy by increasing the size and complexity of a network may be interpreted as the development of less intelligent neural networks. This is supported anecdotally in the literature by the tendency of most neural networks to memorize images with randomized labels during training.\cite{zhang2016understanding}

Inspired by the discovery of the neural efficiency hypothesis from human intelligence research, this work describes a metric for assessing neural efficiency in neural network layers. The artificial intelligence quotient (aIQ) is defined as a combination of neural network efficiency and model performance, so that a neural network with "high intelligence" uses a small number of neurons to make accurate predictions.

\section{Approach}
\subsection{State Space}
Prior attempts to increase the efficiency of a neural network were based on removal of weights (pruning) based on weight magnitude or gradients of weights during backpropagation \cite{NIPS1989_250,NIPS1992_647} or analysis of firing frequency.\cite{hu2016network} In this manuscript, the state space of a single layer is analyzed, where a single state is the collective output of a neural layer given a single set of inputs. Since the output values of all neurons for a given set of inputs are generally passed to the subsequent layer, it may be beneficial to analyze how neurons fire as a collective rather than analyzing individual neurons.

If the output of a neuron layer defines a single state, then the state space is the frequency that each state of a layer occurs as all images in the train or test data pass through the network. When one image is passed through a convolutional neural network, convolutional layers will generate multiple states per image. In contrast, dense layers will generate only one state per image. In this manuscript, neuron outputs are quantized as either firing (output is greater than zero) or non-firing (output is less than or equal to 0). However, even with quantization the state space could still be unmanageably large since the number of possible states in a layer after quantization will be $2^{N_l}$, where $N_l$ is the number of neurons in a layer. For most ANNs, the number of neurons in a layer frequently exceeds 64, meaning most computers would be incapable of creating a memory address for each layer state. In reality it might be expected that significantly fewer states are actually generated, so bins for a layer state are only created when observed.

\subsection{Neural Efficiency}

Neural efficiency is defined here as utilization of state space, and it can be measured by entropic efficiency. If all possible states are recorded for data fed into the network, then the probability, $p$, of a state occuring can be used to calculate Shannon's entropy, $E_l$, of network layer $l$:

$$E_{l} = -\sum p * log_2(p)$$

Intuitively, $E_l$ is an estimation of the minimum number of neurons required to encode the information exported by the neural layer if the output information could be perfectly encoded. The maximum theoretical entropy of the layer will occur when all states occur the same number of times, and the entropy value will be equal to the number of neurons in the layer, $N_l$. Neural efficiency, $\eta_{l}$ can then be defined as the entropy of the observed states relative to the maximum entropy:

$$\eta_{l} = \frac{E_l}{N_l}$$

Thus, neural efficiency, $\eta_{l}$, is defined as state space efficiency using Shannon's entropy with a range of 0-1. Neural efficiency values close to zero are likely to have more neurons than needed to process the information in the layer, while neuron layers with neural efficiency close to one are making maximum usage of the available state space. Alternatively, high neural efficiency could also mean too few neurons are in the layer.

\subsection{Artificial Intelligence Quotient}

Neural efficiency is a characteristic of intelligence, but so is task performance. Therefore, an intelligent algorithm should perform a task with high accuracy and efficiency. Using $\eta_{l}$ as layer efficiency, the neural network efficiency, $\eta_N$, can be calculated as the geometric mean of all layer efficiencies in a network containing $L$ number of layers:

$$\eta_{N} = \Bigl(\prod_{l=1}^{L} \eta_{l}\Bigr)^\frac{1}{L}$$

Then, the artificial intelligence quotient (aIQ) can be defined as:

$$aIQ = \Bigl(P^{\beta}*\eta_{N}\Bigr)^\frac{1}{\beta+1}$$

where $P$ is the performance metric and $\beta$ is a tuning parameter to give more or less weight to performance at the cost of $\eta_N$.

\section{Experiments}

\subsection{Exhaustive LeNet Training}

To evaluate neural efficiency and aIQ, two types of neural networks were trained on the MNIST digits data set.\cite{lecun98} The first network (LeNet-300-100) consists of two densely connected layers followed by a classification layer. The second network (LeNet-5) consisted of two convolutional layers, each followed by a max pooling layer (2x2 pooling with stride 2), and a densely connected layer followed by a classification layer. All layers used exponential linear unit (ELU) activation,\cite{clevert2015fast} L2 weight regularization (0.0005), and no batch normalization or dropout was used. Standard stochastic gradient descent with Nesterov updates was used with a static learning rate of 0.001 and momentum of 0.9. Training was stopped when the training accuracy did not increase within five epochs of a maximum value.

For each neural network architecture, the number of neurons in every layer were varied from 2 to 1024 by powers of 2. All combinations of layer sizes were trained, with eleven replicates using different random seeds to determine variability resulting from different initializations. This resulted in a total of 1,100 different LeNet-300-100 trained networks, and 11,000 different LeNet-5 trained networks. Additional models were trained to identify the specific architecture with the highest aIQ, so that the total number of LeNet-300-100 models was 2,575 and total number of LeNet-5 models was 26,269. Models were constructed and trained using Tensorflow 2.1, and networks were trained in parallel on two gpu servers with 8 NVidia Quadro RTX 8000s each. These networks serve as a baseline for comparison in subsequent experiments.

Once models were trained, the entropy of each neuron layer in a network was calculated based on the distribution of all possible layer states generated by passing all test data through the network (training data was also evaluated separately). Then, aIQ was calculated with $\beta = 2$ to give a nominal preference for higher accuracy networks.

\subsection{Dropout and Batch Normalization}

In the context of neural layer efficiency, batch normalization was hypothesized to be a method to improve efficiency while dropout is a method to decrease efficiency. The rationale for batch normalization improving neural efficiency is that neuron activation is driven toward the center of the distribution of neuron outputs. As the firing frequency of each neuron approaches 50\%, the entropy is more likely to obtain the maximum value. In contrast, dropout was hypothesized to decrease the available state space during training by dropping the outputs of neurons, effectively decreasing the maximum entropy value. This concept is also in line with the original paper that claimed that dropout creates redundancies within the network, and redundancies are innefficient. To test the effect of dropout and batch normalization, neural networks were trained with the same number of neurons as described in the Exhaustive LeNet Training section, except a dropout layer ($p = 50\%$) or batch normalization layer was added after every hidden layer. For networks with batch normalization or dropout layers, only 3 replicates were trained instead of 11.

\subsection{Memorization and Generalization Tests}

If aIQ provides an assessment of capacity to learn general rules rather than memorize training inputs, it might be expected that network architectures with high aIQ perform well on data sets with randomized labels. The reason for this is that for low aIQ networks with low $\eta_N$, training inputs are memorized because the state space is likely much larger than space of observed states. This means new data may be classified correctly or incorrectly based on how similar an image was to one of the memorized inputs. However, for high aIQ networks there is insufficient bandwidth to create a special state for an input with a randomized labe. To test this, network architectures were trained from scratch where 25\%, 50\%, 75\% or 100\% of the training labels were randomized. Then, accuracy and neural efficiency was measured for both the test and train data sets without randomized labels. To test the capacity of high aIQ networks to generalize to new data, trained networks were use to evaluate the EMNIST data set,\cite{cohen2017emnist} which contains 280,000 additional digit images in the same formats as the original 70,000 digit images contained in MNIST. For both memorization and generalization tests, models were trained as prevoiusly described with batch normalization layers but only three replicates were trained per model.

\section{Results}

\subsection{Accuracy, aIQ, and Neural Efficiency}

\begin{figure}[htb!]
  \centering
%  \fbox{\rule[-.2in]{0in}{5.5in} \rule[-.2in]{5.5in}{0in}}
  \includegraphics[width=\linewidth]{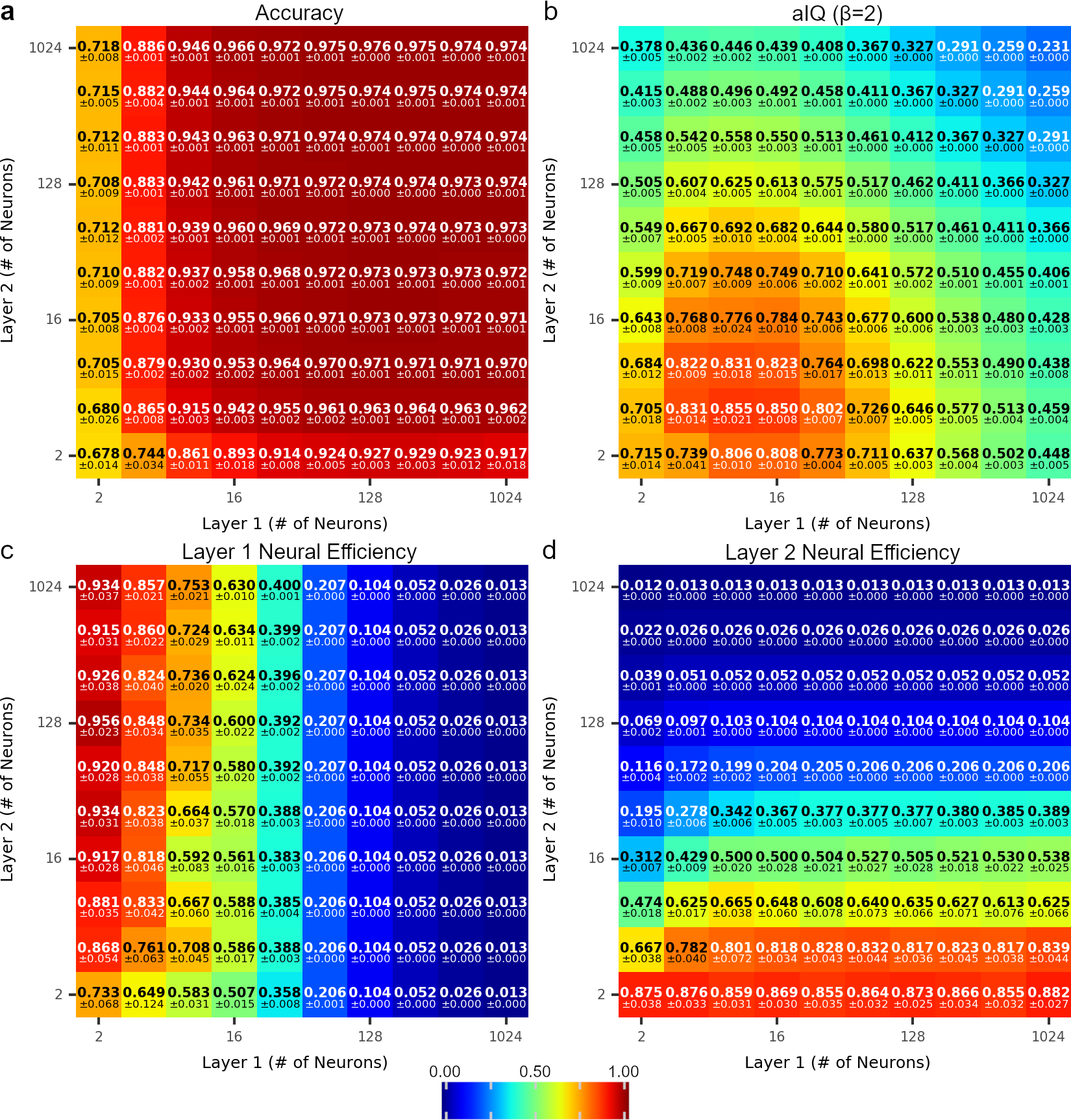}
  \caption{Trends in Accuracy (a), aIQ (b), and Neural Efficiency for hidden layer 1 (c) and hidden layer 2 (d) in the LeNet-300-100 model. Data shown is for test data. All values range from a 0-1, and values are the mean $\pm$ 95\% confidence interval of 11 replicates.}
  \label{fig:01}
\end{figure}

The LeNet-300-100 network permits easy visualization of trends in model accuracy, aIQ, and the efficiency of each layer due to the network only containing two hidden layers. For the training data (not shown), accuracy increased monotonically with the number of neurons added in each layer, but the test data showed a slight decrease in accuracy as the number of neurons in hidden layer 1 ($N_{1}$) contained more than 128 neurons (Figure \ref{fig:01}a). In contrast, aIQ ($\beta=2$) values reached a local maximum when $N_1 = 8$  and $N_2 = 4$ (Figure \ref{fig:01}b). The decrease in aIQ values is due to the trend in neural efficiency to decrease as the number of neurons in a layer increases (Figure \ref{fig:01}c-d). While it may be expected that decreasing the number of neurons in a layer would increase the neural efficiency, it was unexpected how strong of an impact changing the number of neurons in other layers could impact neural efficiency. For example, for networks where $N_{1} = 8$, the neural efficiency of layer 1, $\eta_1$, generally increased as $N_2$ increased, except there was a local minimum at $N_2 = 16$. Local minima can be observed in other areas where the number of neurons in either layer is held constant and observing the changes in efficiency for the same layer (e.g. when $N_2 = 8$, a local minima occurs when $N_1 = 32$). Similar trends were observed in the LeNet-5 models, where changes in the number of neurons in one layer affected the efficiency of other layers.

\begin{table}[htb!]
  \caption{The top three network architectures by accuracy and aIQ.}
  \label{table:01}
  \centering
  \begin{tabular}{cccccc}
    \toprule
    \multicolumn{3}{c}{LeNet-300-100 ($N_{l}$)}         \\
    \cmidrule(r){1-3}
    Layer 1 & Layer 2 && Accuracy (\%)$^{\dagger}$ & aIQ ($\beta=2$)$^{\dagger}$ & Parameters (fold decrease)  \\
    \midrule
    128  & 1024 & & \textbf{97.58 $\pm$ 0.04} & 32.70 $\pm$ 0.01 & 242,826 (1x)    \\
    256  & 1024 & & 97.54 $\pm$ 0.06 & 29.12 $\pm$ 0.01 & 474,378 (0.5x)   \\
    64    & 1024 & & 97.53 $\pm$ 0.08 & 36.67 $\pm$ 0.02 & 127,050 (1.9x)   \\
    \addlinespace[0.25em]
    11    & 4       & & 92.91 $\pm$ 0.19 & \textbf{86.41 $\pm$ 0.71} & 8,733 (27.8x)    \\
    11    & 5       & & 93.59 $\pm$ 0.29 & 85.93 $\pm$ 1.04 & 8,755 (27.7x)   \\
    7      & 4       & & 90.76 $\pm$ 0.25 & 85.90 $\pm$ 1.00 & 5,577 (43.5x)   \\
    \midrule
    \multicolumn{3}{c}{LeNet-5 ($N_{l}$)}         \\
    \cmidrule(r){1-3}
    Layer 1 & Layer 2 & Layer 3 & Accuracy (\%)$^{\dagger}$ & aIQ ($\beta=2$)$^{\dagger}$ & Parameters (fold decrease)  \\
    \midrule
    1024 & 1024 & 1024 & \textbf{99.16 $\pm$ 0.02} & 24.28 $\pm$ 0.004 & 43,030,538 (1x)    \\
    1024 & 128   & 512   & 99.15 $\pm$ 0.03 & 33.02 $\pm$ 0.008 & 4,357,770 (9.9x)    \\
    1024 & 512   & 1024 & 99.14 $\pm$ 0.03 & 26.22 $\pm$ 0.005 & 21,534,218 (2.0x)  \\
    \addlinespace[0.25em]
    3       & 9       & 4       & 96.84 $\pm$ 0.18 & \textbf{88.28 $\pm$ 0.46} & 1,392 (30,912.9x)  \\
    3       & 5       & 5       & 96.72 $\pm$ 0.16 & 88.18 $\pm$ 1.52 & 923 (46,620.3x)  \\
    3       & 4       & 4       & 95.37 $\pm$ 0.19 & 88.02 $\pm$ 0.64 & 692 (62,182.9x)  \\
    \bottomrule
    \multicolumn{6}{l}{$\dagger$ Accuracy and aIQ values are mean $\pm$ 95\% CI (n=11). aIQ values are x100.} \\
    \multicolumn{6}{l}{Data shown is for metrics calculated on the test data set.} \\
  \end{tabular}
\end{table}

Additional networks were trained to identify networks with the highest aIQ for each architecture. Analysis of the top three neural networks for accuracy or aIQ for both LeNet models are shown in Table \ref{table:01}. For the LeNet-300-100 models, the model with the highest test accuracy ($N_1 = 128$, $N_2 = 1024$) achieved an accuracy of 97.58\% $\pm$ 0.04\% with an aIQ of 32.7 $\pm$ 0.01 (values are mean $\pm$ 95\% CI, n=11). The highest aIQ model ($N_1 = 11$, $N_2 = 4$) had an accuracy of 92.91\% $\pm$ 0.19\% with an aIQ of 86.41 $\pm$ 0.71. Thus, the highest aIQ network was 4.76\% less accurate but contained ~27.8 times fewer parameters.

The differences between the highest accuracy and highest aIQ networks were even more drastic for the LeNet-5 models. The highest accuracy network ($N_1 = N_2 = N_3 = 1024$) had an accuracy of 99.58\% $\pm$ 0.02\% and an aIQ of 24.28 $\pm$ 0.004 (values are mean $\pm$ 95\% CI, n=11). However, the highest aIQ network ($N_1 = 3$, $N_2 = 9$, $N_3 = 4$) had an accuracy of 96.84\% $\pm$ 0.18\% and aIQ of 88.28 $\pm$ 0.46. The highest aIQ network had a lower accuracy by 2.32\% but contained 30,912.9 times fewer parameters.

\subsection{Batch Normalization and Dropout as Neural Efficiency Modifiers}

\subsubsection{Batch Normalization}

Batch normalization generally increased the accuracy and $\eta_N$ for both LeNet-300-100 and LeNet-5 networks, resulting in a rise in aIQ for most networks (Table \ref{table:02}). For LeNet-300-100 networks, 59.26\% of network architectures with batch normalization had a mean $\eta_N$ (n=3 replicates) higher than the mean $\eta_N$ of corresponding networks trained without batch normalization (n=11). The reason why all networks with batch normalization do not have higher efficiency than networks without batch normalization can be explained by state space limits. For fully connected layers, the largest number of observable states would be equal to the number of inputs. This was confirmed by looking at the state space of the last dense layer to verify only 60,000 states (i.e. the number of training examples) were observed for large neuron layers when evaluating the training data. Thus, batch normalization could not increase the entropy of large layers. When only small networks are considered ($N_1 \leq 16$, $N_2 \leq 16$), 74.13\% of networks with batch normalization had a mean $\eta_N$ higher than the same network trained without batch normalization. In addition to higher network efficiency, neural networks trained with batch normalization also had higher accuracy on test data, where 77.78\% of networks with batch normalization achieved higher accuracies than the same networks without batch normalization. Since both accuracy and $\eta_N$ increase with batch normalization, it is unsurprising that aIQ increased in 74.07\% of networks where all layers had 16 or fewer neurons.

For LeNet-5 networks, 74.2\% of all networks with batch normalization had a mean $\eta_N$ (n=3) higher than corresponding networks without batch normalization (n=11). However, the accuracy for networks with batch normalization caused the accuracy of networks to decrease on average, with only 32.60\% of networks achieving a higher accuracy than corresponding networks without batch normalization. When considering only small networks where $N_l \leq 16$ for all layers, 92.6\% of networks with batch normalization had a higher accuracy. This discrepancy can be explained by batch normalization increasing the likelihood of large networks memorizing training inputs, leading to the worse performance on the test data (i.e. overtraining). For $\eta_N$, 71.10\% of all networks with batch normalization had a higher $\eta_N$ relative to their corresponding networks without batch normalization, and $\eta_N$ decreased to 64.35\% for networks where all layers had less than 16 neurons. Overall, 74.20\% of all networks had a higher aIQ when trained with batch normalization.

These results confirm the hypothesis that batch normalization generally acts to improve $\eta_N$ in addition to improving classification accuracy (Table \ref{table:02}), making neural networks "more intelligent" as assessed by aIQ.

\begin{table}[htb!]
  \caption{The top network architectures by aIQ when batch normalization (BatchNorm), drop out (Dropout), or neither (None) are added to every layer.}
  \label{table:02}
  \centering
  \begin{tabular}{ccccccc}
    \toprule
    \multicolumn{3}{c}{LeNet-300-100 ($N_{l}$)}         \\
    \cmidrule(r){1-3}
    Layer 1 & Layer 2 && Modifier & Accuracy (\%)$^{\dagger}$ & aIQ ($\beta=2$)$^{\dagger}$ & $\eta_N$ (\%)  \\
    \midrule
    11    & 4       && \textbf{None} & 92.91 $\pm$ 0.19 & \textbf{86.41 $\pm$ 0.71} & \textbf{74.77 $\pm$ 1.72}    \\
            &          && BatchNorm & \textbf{93.50 $\pm$ 0.36} & 82.10 $\pm$ 1.44 & 63.33 $\pm$ 3.07    \\
            &          && Dropout     & 78.41 $\pm$ 3.82 & 74.36 $\pm$ 4.06 & 66.89 $\pm$ 4.47    \\
    \addlinespace[0.25em]
    10    & 6       && None          & 93.52 $\pm$ 0.14 & 83.16 $\pm$ 2.52 & 66.20 $\pm$ 5.75    \\
            &          && \textbf{BatchNorm} & \textbf{93.65 $\pm$ 0.44} & \textbf{87.76 $\pm$ 0.44} & \textbf{77.07 $\pm$ 1.16}    \\
            &          && Dropout     & 83.63 $\pm$ 0.63 & 75.62 $\pm$ 0.93 & 61.83 $\pm$ 1.49    \\
    \addlinespace[0.25em]
    7      & 5       && None          & 91.53 $\pm$ 0.31 & 83.88 $\pm$ 1.75 & 70.65 $\pm$ 4.21    \\
            &          && BatchNorm &\textbf{ 91.91 $\pm$ 0.25} & \textbf{86.42 $\pm$ 1.20} & \textbf{76.43 $\pm$ 2.76}    \\
            &          && \textbf{Dropout}     & 80.63 $\pm$ 2.96 & 78.31 $\pm$ 2.44 & 73.88 $\pm$ 1.47    \\
    \midrule
    \multicolumn{3}{c}{LeNet-5 ($N_{l}$)}         \\
    \cmidrule(r){1-3}
    Layer 1 & Layer 2 & Layer 3 & Modifier & Accuracy (\%)$^{\dagger}$ & aIQ ($\beta=2$)$^{\dagger}$ & $\eta_N$ (\%)  \\
    \midrule
    3      & 9       & 4             &\textbf{ None } & 96.84 $\pm$ 0.18 & \textbf{88.28 $\pm$ 0.46} & 73.38 $\pm$ 1.22    \\
            &          &                & BatchNorm &\textbf{ 96.95 $\pm$ 0.58} & 82.44 $\pm$ 3.08 & 59.73 $\pm$ 6.47    \\
            &          &                & Dropout      & 79.72 $\pm$ 4.96 & 77.82 $\pm$ 2.91 & \textbf{74.23 $\pm$ 0.99}    \\
    \addlinespace[0.25em]
    2      & 8       & 8             & None          & 97.83 $\pm$ 0.10 & 85.58 $\pm$ 2.28 & 65.81 $\pm$ 5.03    \\
            &          &                & \textbf{BatchNorm} & \textbf{98.05 $\pm$ 0.20} & \textbf{88.12 $\pm$ 1.68} & 71.21 $\pm$ 4.30    \\
            &          &                & Dropout     & 88.06 $\pm$ 2.00 & 82.75 $\pm$ 0.38 & \textbf{73.16 $\pm$ 4.21}    \\
    \addlinespace[0.25em]
    3      & 4       & 7             & None          & 97.17 $\pm$ 0.22 & 83.71 $\pm$ 2.65 & 62.55 $\pm$ 5.74    \\
            &          &                & BatchNorm &\textbf{ 97.27 $\pm$ 0.34} & 84.91 $\pm$ 2.16 & 64.76 $\pm$ 4.45    \\
            &          &                & \textbf{Dropout}     & 90.15 $\pm$ 1.66 & \textbf{85.45 $\pm$ 1.47} & \textbf{76.82 $\pm$ 3.32}    \\
    \bottomrule
    \multicolumn{7}{l}{$\dagger$ Accuracy, aIQ, and Efficiency values are mean $\pm$ 95\% CI (n=11 for None, n=3 otherwise).} \\
    \multicolumn{7}{l}{aIQ values are x100. Data shown is for metrics calculated on the test data set.} \\
  \end{tabular}
\end{table}

\subsubsection{Dropout}

Dropout had different effects on the LeNet-300-100 and LeNet-5 architectures. Dropout generally decreased $\eta_N$ in LeNet-300-100 networks and increased $\eta_N$ in the LeNet-5 networks, but decreased aIQ in nearly all networks due to a drop in accuracy for nearly all networks (Table \ref{table:02}). For LeNet-300-100 networks, none of the networks trained with dropout had an accuracy higher than corresponding networks trained without dropout, and only 5.59\% of networks had a higher efficiency. Accordingly, no networks with dropout had a higher aIQ relative to corresponding networks without dropout.

The trends were surprisingly different for LeNet-5 networks. While none of the networks with dropout had an accuracy higher than corresponding networks without dropout, 60.60\% of dropout networks had a higher $\eta_N$. When considering neural networks with 16 or fewer neurons in each layer, 99.33\% of networks with dropout had a higher $\eta_N$ than their corresponding networks without dropout. However, only 15.60\% of dropout networks had a higher aIQ than corresponding networks without dropout, meaning the increase in efficiency was not sufficient to offset the decrease in accuracy in the majority of networks.

The general conclusion from these results is that dropout generally decreases accuracy, leading to a drop in aIQ. The reason why accuracy dropped in all networks may be due to a variety of factors, including the dropout rate being too high ($p = 50\%$) or because dropout was placed in every layer of the network. The explanation for why dropout appeared to decrease $\eta_N$ in LeNet-300-100 networks but increased $\eta_N$ in LeNet-5 networks may be due to LeNet-300-100 only having dense layers while LeNet-5 contained convolutional layers. It is known that dropout can have a nominal or detrimental impact on network performance because dropout layers add noise to the network\cite{park2017}. As a result, the increased $\eta_N$ may be due to increased noise in the convolutional layers leading to higher entropy. A better assessment of the effect of dropout for convolutional layers may be to use dropout layers with different dropout rates or use dropout layer types constructed specifically for convolutional layers, such as dropblock or spatial dropout (a survey of different dropout types was performed by Labach \textit{et al}).\cite{labach2019survey}

\subsection{Memorization and Generalization}

\begin{table}[htb!]
  \caption{Memorization tests on the networks with the highest aIQ or accuracy.}
  \label{table:03}
  \centering
  \begin{tabular}{cccccccc}
    \toprule
    \multicolumn{3}{c}{LeNet-300-100 ($N_{l}$)} & \multicolumn{5}{c}{Accuracy (\%)}         \\
    \cmidrule(r){1-3}
    \cmidrule(r){4-8}
    Layer 1 & Layer 2 && 0\% Rand & 25\% Rand & 50\% Rand & 75\% Rand & 100\% Rand  \\
    \midrule
    1024& 1024 && 98.03 & 85.91 & 63.43 & 36.99 & 9.48 \\
    \addlinespace[0.25em]
    10    & 6       && 93.65 & 91.56 & 89.59 & 84.68 & 10.89 \\
    \midrule
    \multicolumn{3}{c}{LeNet-5 ($N_{l}$)}         \\
    \cmidrule(r){1-3}
    Layer 1 & Layer 2 & Layer 3 & 0\% Rand & 25\% Rand & 50\% Rand & 75\% Rand & 100\% Rand  \\
    \midrule
    1024& 1024 &   1024     & 99.11 & 93.99 & 76.07 & 44.24 & 10.34   \\
    \addlinespace[0.25em]
    2      & 8       &   8           & 98.05 & 96.73 & 95.69 & 92.51 & 7.49   \\
    \bottomrule
    \multicolumn{8}{l}{The \% Rand indicates the percentage of labels that were randomized prior to training.} \\
  \end{tabular}
\end{table}

\begin{table}[htb!]
  \caption{Generalization tests on the networks with the highest aIQ or accuracy.}
  \label{table:04}
  \centering
  \begin{tabular}{ccccccc}
    \toprule
    \multicolumn{3}{c}{LeNet-300-100 ($N_{l}$)} & \multicolumn{2}{c}{MNIST Accuracy (\%)} & \multicolumn{2}{c}{EMNIST Accuracy (\%)}        \\
    \cmidrule(r){1-3}
    \cmidrule(r){4-5}
    \cmidrule(r){6-7}
    Layer 1 & Layer 2 && 0\% Rand & 75\% Rand & 0\% Rand & 75\% Rand \\
    \midrule
    1024& 1024 && 98.03 & 36.99 & 76.21 & 24.93 \\
    \addlinespace[0.25em]
    10    & 6       && 93.65 & 84.68 & 58.43 & 55.46 \\
    \midrule
    \multicolumn{3}{c}{LeNet-5 ($N_{l}$)}         \\
    \cmidrule(r){1-3}
    Layer 1 & Layer 2 & Layer 3 & 0\% Rand & 75\% Rand & 0\% Rand & 75\% Rand \\
    \midrule
    1024& 1024 &   1024     & 99.11 & 44.24 & 90.56 & 34.16   \\
    \addlinespace[0.25em]
    2      & 8       &   8           & 98.05 & 92.51 & 89.36 & 74.67   \\
    \bottomrule
    \multicolumn{7}{l}{The \% Rand indicates the percentage of labels that were randomized prior to training on MNIST.} \\
  \end{tabular}
\end{table}

To test the resistance of networks to overfitting/memorization, a percentage of image labels were randomized before training as previously described by Zhang \textit{et al}.\cite{zhang2016understanding} Overfitting occurs when incorrect, random labels are learned and is analagous to memorizing the label for an image. Table \ref{table:03} shows the results for the neural networks with the highest aIQ and the highest accuracy from the batch normalization tests for both the LeNet-300-100 and LeNet-5 networks. When none of the labels are randomized (0\%), the accuracy of the largest network is higher than the best aIQ network for both LeNet-300-100 and LeNet-5. When 25\%-75\% of the labels were randomized, the network with the highest aIQ had the best accuracy. Even when 75\% of the labels were randomly assigned, the high aIQ networks were able to achieve accuracies of 84.68\% and 92.51\% for the LeNet-300-100 and LeNet-5 networks, respectively. These numbers are considerably better than the neural networks with the highest accuracies in the batch normalization tests, which had test accuracies of 36.99\% and 44.24\% when trained on data with 75\% of the labels randomized. Both types of networks had bad performance regardless of original aIQ or accuracy values when all labels were randomly assigned (100\%). This result demonstrates that high aIQ networks have the property of being resistant to overtraining and memorization, learning the correct classification weights even when a majority of the input labels are incorrect.

To test the capacity of networks to generalize results to a larger, more diverse data set, the highest aIQ or accuracy networks were trained on the MNIST data set and then classification accuracy was measured on the EMNIST data set. The EMNIST digits data set has a similar format to MNIST, except it contains 280,000 more samples.\cite{cohen2017emnist} The general trend was that the highest accuracy network performed better on EMNIST than MNIST, with significant differences between the LeNet-300-100 and LeNet-5 networks (Table \ref{table:04}). The LeNet-300-100 network with the highest accuracy on MNIST (98.03\%) was much better than the accuracy on EMNIST (76.21\%), while the highest aIQ network had a much larger decrease in accuracy from MNIST to EMNIST (93.65\% to 58.43\% respectively). In contrast, the differences between the highest accuracy and aIQ networks were less drastic for the LeNet-5 networks. The EMNIST accuracy for the highest accuracy network was 90.56\% and was 89.36\% for the highest aIQ network. These experiments were repeated after training the same networks on MNIST with 75\% of the labels randomized, and as expected the highest accuracy networks had a significant decrease in performance on EMNIST while the highest aIQ networks performed considerably better for both LeNet-300-100 and LeNet-5 networks. This data demonstrates that high aIQ convolutional neural networks do not considerably underperform in comparison to much larger networks when performance is assessed on a much larger, diverse data set, but dense networks with high aIQ may not generalize as well.

\section{Discussion}

The major contribution of this work is the establishment of neuron layer state space and the capacity to use state space as a means of evaluating neuron layer utilization. Quantizing neurons into on/off positions that encode information collectively as a neural layer state provides a different perspective on how data is processed in the network. One prevailing thought is that neurons are discrete units that encode individual features, but the work here suggests that the information an individual neuron encodes may have value in the context of the other neurons it fires with. One advantage of conceptualizing the flow of information between layers using state space is the number of tools that become available for network analysis. In this work, a rudimentary metric was created to understand layer utilization, but many other methods of analyzing state space could be used such as relative entropy of a single layer or the mutual information between two layers. Further investigation of state space may help to further compress the size of the network without significant decrease in accuracy, and may even permit training the number of neurons in a layer. While neural architecture search has become a topic of interest to search for an ideal network computational cell, few if any of these methods include parameters to learn layer sizes.

There are a few deficits in the current approach that should be resolved in future work. First is the issue of class imbalances. If there are class imbalances in either the training or test data, it would be expected that the efficiency metrics would be skewed. Class imbalances in the training data might cause more neurons in the network to be dedicated to classification of the most common class while imbalances in the test data might over represent states that occur. A second issue is the underlying assumption that maximum efficiency is achieved when all states occur at the same frequency. It might be expected that some states occur far more frequently than others so that an ideal distribution of states might look more like an exponential distribution. This was superficially confirmed by looking at the distribution of states in the LeNet-300-100 networks by analyzing networks with $N_1 = 8$ around the local minimum when $N_2 = 16$ (see Figure \ref{fig:01}). Therefore, some other metric of calculating efficiency that accounts for an ideal distribution of states might be a better measure of efficiency.

A third issue is implementation. Recording and processing data collected in state space is expensive. For some networks, the number of neurons in a convolutional layer can be 128 or more, meaning that at least $2^{128}$ layer states are possible and the states for every location in an image should be tracked to calculate the entropy. The current implementation of calculating entropy is not practical for larger networks that contain many more layers and layers with larger numbers of neurons (such as AlexNet with 4,096 neurons in the dense layers).\cite{alex2012} One potential solution to this might be to create a method of approximating the entropy by collecting sufficient information on a layer by layer basis to capture the shape of the distribution rather than recording every observed state.

Finally, one topic not investigated here is how data augmentation impacts efficiency and aIQ. Data augmentation is generally used to help improve accuracy and generalization, likely because it helps to mitigate memorization. Networks with high aIQ were small and were fairly resistant to memorization (see Table \ref{table:03}), therefore it might be expected that certain types of augmentation (i.e. random cropping) might not improve performance when training a high aIQ network, but other types of augmentation might help (i.e. image flipping).

\section{Conclusions}

This work introduced the concept of state space, and demonstrated how analysis of state space is a useful tool for assessing neural layer efficiency. High aIQ networks were shown to have desireable properties, such as resistance to overtraining and comparable general performance on EMNIST to much larger networks. Future work with state space should establish better metrics of layer efficiency and methods of computing efficiency and evaluation of larger networks than the two models presented in this paper. State space may provide insight into sizing neural network layers to vastly decrease the size of existing neural networks.

\newpage

\section*{Broader Impact}

One benefit of understanding neural networks in terms of state space is that guidelines on how many neurons to place in a layer can be established knowing only superficial information about the training data. The current thought is that more neurons leads to higher accuracy, but this does not appear to provide improved generalization for the small convolutional models tested here (see Table \ref{table:04}). Using the concept of state space, the number of states in a dense layers cannot exceed the number of input images. If there are X training examples, then $ceil(log_{2}X)$ neurons are sufficient to memorize every training image. As an example, the ImageNet data set has ~14 million images (leading to \textasciitilde2\textsuperscript{24} states) which is considerably smaller than the available statespace of AlexNet's dense layers with 4096 neurons.\cite{alex2012} This means that the statespace of AlexNet is \textasciitilde6*10\textsuperscript{1225} times larger than the number of available training examples in ImageNet. Due to random initialization, it is doubtful that a dense layer would memorize every training image if it contains exactly the number of neurons required to memorize all inputs. However, assuming the network is generating general rules for classification, the entire bandwidth of the channel should never need to be used. An analagous guideline could be applied to convolutional layers, where all combinations of pixel intensities of an 8-bit, grayscale image in a 3x3 grid could be perfectly represented by 72 neurons ($256^9 = 2^{72}$), so the first convolutional layer of a network should never contain more than 72 neurons when analyzing 8-bit grayscale images. Thus, simple upper limits on the number of neurons in different layers can be inferred from the implications of state space, where the upper limits may be considerably smaller than the number of neurons are currently observed in some networks such as AlexNet.

Using the guidelines laid out above, it is reasonable to say that most networks that are created contain many more neurons than needed and may explain why most neural networks are prone to attack vectors. For example, adversarial networks can be trained to add imperceptable amounts of noise to an image to cause the neural network to misclassify the image.\cite{goodfellow2014explaining} It is plausible that these attack vectors take advantage of noise in an overparameterized network, a problem that a smaller neurals network may not face. Thus, use of state space and neural efficiency may help to make networks more resistant to such attacks.

\section*{Acknowledgements and Disclosure of Funding}

\bibliographystyle{unsrt}
\bibliography{references}

\end{document}